\documentclass[a4paper]{article}

\usepackage[T1]{fontenc}
\usepackage[utf8x]{inputenc}
\usepackage[english]{babel}
\usepackage[colorlinks=true, allcolors=blue]{hyperref}

\newcommand{\email}[1]{\href{mailto:#1}{\tt{\nolinkurl{#1}}}}
\newcommand{\orcid}[1]{ORCID: \href{https://orcid.org/#1}{\tt{\nolinkurl{#1}}}}

\usepackage[sfdefault,lf]{carlito}
\usepackage[parfill]{parskip}

\usepackage{fancyhdr}
\usepackage{authblk}
\usepackage[a4paper,top=3cm,bottom=2cm,left=3cm,right=3cm,marginparwidth=1.75cm]{geometry}

\usepackage{amssymb}
\usepackage{amsmath}
\usepackage{graphicx}
\usepackage{booktabs}
\usepackage{color}
\usepackage{float}
\usepackage{soul}

\usepackage[colorinlistoftodos]{todonotes}
\usepackage[version=3]{mhchem}
\usepackage{mathtools} 

\title{Data-efficient and interpretable inverse materials design using a disentangled variational autoencoder}
\author[1, 2, $\dag$, *]{Cheng Zeng}
\author[1, $\dag$, *]{Zulqarnain Khan}
\author[2]{Nathan L. Post}
\affil[1]{Institute for Experiential AI, Northeastern University, Boston, MA 02115, United States}
\affil[2]{The Roux Institute, Northeastern University, Portland, ME 04101, United States}
\affil[$\dag$]{These authors contribute equally: Cheng Zeng, Zulqarnain Khan.}
\affil[*]{Corresponding author: \email{c.zeng@northeastern.edu}, \email{z.khan@northeastern.edu}}
\date{}

\begin{document}
\maketitle
\thispagestyle{plain}

\begin{abstract}
Inverse materials design has proven successful in accelerating novel material discovery. Many inverse materials design methods use unsupervised learning where a latent space is learned to offer a compact description of materials representations. A latent space learned this way is likely to be entangled, in terms of the target property and other properties of the materials. This makes the inverse design process ambiguous. Here, we present a semi-supervised learning approach based on a disentangled variational autoencoder to learn a probabilistic relationship between features, latent variables and target properties. This approach is data efficient because it combines all labelled and unlabelled data in a coherent manner, and it uses expert-informed prior distributions to improve model robustness even with limited labelled data. It is in essence interpretable, as the learnable target property is disentangled out of the other properties of the materials, and an extra layer of interpretability can be provided by a post-hoc analysis of the classification head of the model. We demonstrate this new approach on an experimental high-entropy alloy dataset with chemical compositions as input and single-phase formation as the single target property.
High-entropy alloys were chosen as example materials because of the vast chemical space of their possible combinations of compositions and atomic configurations.
While single property is used in this work, the disentangled model can be extended to customize for inverse design of materials with multiple target properties.
\end{abstract}

\noindent\textbf{Keywords}: inverse materials design; high-entropy alloys; disentangled variational autoencoder; interpretable methods

\section{Introduction}

Materials play a pivotal role in shaping the modern society and many grand technological challenges are materials challenges. These range from lower-cost battery materials for energy storage, to quantum computing materials and bio-compatible materials for healthcare applications~\cite{li2017,de2021,tahmasebi2020}. Thanks to advances in high-throughput computing~\cite{merchant2023}, robotics~\cite{szymanski2023}, machine learning (ML) force fields~\cite{merchant2023} and open data repositories of materials~\cite{jain2013, choudhary2020}, materials design and discovery have now reached an unprecedented rate and scale~\cite{knapp2022}. Although advances in algorithm and hardware significantly reduce the computation time to evaluate each material candidate, material scientists still need to often scan a wide range of materials candidates to pinpoint a small number of potential materials with desired properties~\cite{zeng2024}.

Inverse materials design unlocks the potential to optimize new materials towards a target property. In general there are four approaches for inverse materials design~\cite{wang2022}; high-throughput virtual screening~\cite{afzal2019}, global optimization~\cite{geng2019}, reinforcement learning~\cite{xian2024} and generative models~\cite{ma2019, popova2018}. Among these, generative models can be very data-efficient as they allow for encoding expert informed information into the model, thereby reducing the amount of data that is required to learn a compact low-dimensional representation. Moreover, the learned representation space can generate new data using the knowledge encoded during training. Comparing various generative models, a study by T\"{u}rk et al showed that variational autoencoder (VAE) is more robust than reinforcement learning and generative adversarial networks because VAE has a better representation of underlying distributions and training a VAE model is easier~\cite{turk2022}.
However, current generative models are primarily deployed as an exploratory unsupervised learning approach to learn a latent representation of materials~\cite{chen2020, wang2022, rao2022}. In the absence of any imposed structure, this latent space learns a representation that entangles all the hidden factors that are needed to successfully generate observed materials and is by design not set-up to learn a disentangled representation where the target property can be separated out from other latent factors.  
Unsupervised learning is not ideal as one will need to perform post-optimization to explore materials with better target properties and may even fail to find any useful materials.
In practice, materials must meet multiple requirements to be suitable for operations.
For instance, materials for aerospace applications must be strong, light-weight and resistant to radiation.
A recent work by Xie and Tomioka et al introduced diffusion-based generative processes together with a fine tuning process to discover materials with multiple targets including magnetic properties and supply chain risk~\cite{zeni2024}.
Although a state-of-the-art discovery rate and materials stability were reported, the full periodicity of the crystal structures restricts the design space to inorganic materials.
The complexity of large datasets and diffusion models hinder a wider application and the interpretability of the methods.
Moreover, the practicality of the new materials discovered in this process is doubtful without expert insight~\cite{cheetham2024}.

Therefore, there is an urgent need to create a workflow for inverse materials design that is data efficient, interpretable, and geared towards multi-property optimization.
The objective of this work is to develop high-entropy alloys (HEAs) which tend to form a single-phase structure.
High-entropy alloys are emerging materials with superior mechanical and functional properties~\cite{xin2020,fu2021,george2020}.
However, high-entropy alloys---defined as alloys with at least four principal elements---live in high-dimensional design space, making it almost impossible to navigate through brute-force experimental or \textit{ab initio} screening.
For example, for an HEA with five possible elements with each element having an integer composition between 5\% and 35\%, the number of possible compositions is on the order of 10$^7$, let alone the massive numbers of possible atomic arrangement for each composition.
Conventional methods to predict single-phase alloys rely on sophisticated design of experiments, thermodynamic modeling and first-principles calculations. These methods are not efficient for optimizing design of high-entropy alloys  because the number of points to search grow combinatorially with the increase of elements~\cite{yao2017, feng2021}.
Although machine learning algorithms have proven to speed up the search of corrosion-resistance or mechanically strong HEAs~\cite{zeng2024, yan2021, tandoc2023}, this forward design strategy still involves scanning a wide configuration and composition space.
Inverse design is theorized to be a much more robust approach as it learns a probabilistic relationship between materials representation and a compact latent space where new materials candidates can be generated from this relationship.
Coupled with an uncertainty estimate, search directions for new materials can be identified with a risk quantification~\cite{rao2022}.
Here we introduce a disentangled generative model using a semi-supervised variational autoencoder for the inverse design of complex materials.
We demonstrate this approach to identify single-phase high-entropy alloy candidates.
Although our demonstration focuses on a single materials property (single phase formation), the approach can also be applied to materials discovery with other target properties including multi-objective targets.
This approach can utilize both labelled and unlabelled data sets. The proposed method is highly versatile and robust, and can be readily extended to other engineering domains where there exists a similar factorizable  input-output relationship.

\section{Methods \& Theories}
\subsection{Dataset}

\begin{figure}[H]
\centering
\includegraphics[width=5.in]{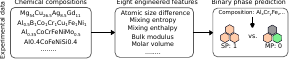}
\caption{Experimental high-entropy alloy dataset for single phase formation.
\label{fig:dataset}}
\end{figure}

This work uses an experimental dataset where inputs are chemical compositions of alloys and outputs are binary phase predictions indicating the formation of single phase (\textbf{SP}: 1) versus multiple phases (\textbf{MP}: 0).
A summary of this dataset is shown in Figure~\ref{fig:dataset}.
It is an up-to-date and well-recognized dataset for HEA single phase prediction collected by Yan et al~\cite{yan2021}.
Empirical rules suggest that the single phase formation relies more on the synergistic effects of mixing different alloys rather than the independent attributes of each element~\cite{yan2021, pei2020}.
Therefore, we also extract eight engineered features from the element-wise compositions.
These eight features include bulk modulus ($k$), molar volume ($V_m$), melting temperature ($T_m$), valence electron concentration ($\mathrm{VEC}$), atomic size difference ($\delta$), Pauling electronegativity difference ($\delta_\chi$), mixing entropy ($\Delta S_{\mathrm{mix}}$) and mixing enthalpy ($\Delta H_{\mathrm{mix}}$). Our proposed model will use both compositional and hand-engineered representations. For simplicity, in the compositional representation, we restrict the element list to the top-30 most frequent elements in the data set i.e. each alloy is represented by a 30-element composition feature vector.
Methods to calculate these features can be found in the supporting information of our previous work~\cite{zeng2024}.
These eight features were found to be informative for the prediction of single phase formation~\cite{zeng2024, yan2021}.
This feature engineering provides a general and compact representation of alloys comprising various elements, offering more generalizability for alloy design.
Given that an experimental data set is used, the trained ML models can thus give predictions that are more likely to be manufactured in practice, without explicitly considering the manufacturing process and environmental conditions.
For the purpose of inverse materials design, we also created another data set where chemical compositions were transformed to element-wise compositions.

\subsection{Proposed Model: Disentangled VAE}
We propose using a disentangled variational autoencoder (DVAE) to predict phase formation and at the same time learn a latent representation where the phase formation is disentangled out to be able to explore other properties of the data. The model is composed of a generative part and a recognition part (Figure~\ref{fig:vae_model}). We describe the model and its training in detail below.
\subsubsection{Generative model}

\begin{figure}[H]
\centering
\includegraphics[width=2.2in]{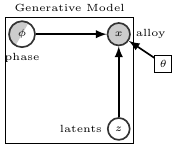}
\includegraphics[width=2.5in]{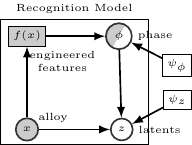}
\caption{Generative (Left) and recognition model (Right) in the disentangled variational autoencoder for inverse design of single-phase high-entropy alloys.}
\label{fig:vae_model}
\end{figure}

Assume $x \in \mathbb{R}^+$ represents an alloy and that each alloy has a phase $\phi \in \{0, 1\}$ that is our target property. The phase is assumed known for at least a subset of the training data. In addition assume $z \in \mathbb{R}^l$ as an  $l$- dimensional latent variable, represents any other factors that may be responsible for generating alloys. With this notation in place, we write the generative model (Figure ~\ref{fig:vae_model}, left) for the data in the form of the following joint probability distribution:
\begin{equation}
    p_\theta(x, \phi, z) = p_\theta(x| \phi, z)p(\phi)p(z)
\end{equation}
Where $\theta$ parameterizes the likelihood using a neural network that takes as input the sampled instances of $\phi$ and $z$. Different choices of priors can be experimented with for each variable, we propose one possible set of choices that best suit the data-space that each variable belongs to. The likelihood of the compositional data is modeled as a multinomial parameterized by the neural network $\theta$ :
\begin{equation}
    x \sim Multinomial(\theta(\phi, z))
\end{equation}

Priors for the binary Phase $\phi$ are modeled using Bernoulli distribution and lastly, we assume a standard normal prior over the latent variable z as a relatively uninformative prior.

\begin{equation}
    \phi \sim Bernoulli(r), z \sim Normal(0, I)
\end{equation}

Note that informed choices for the parameters of these prior distributions can be made, e.g. the phase prior can be chosen to bias the model towards observing more multiphase alloys vs single phase alloys.

\subsubsection{Recognition model}

With the above generative model in place, we can now define the recognition model to perform inference. There are different acceptable recognition models given the above generative model. We propose one that conforms best to our goals of enabling prediction, interpretation, and exploration. The recognition model $q_\psi(\phi, z | x)$  takes as input the data and maps it to the latent representation space, serving as the variational approximation to the otherwise intractable posterior distribution $p(\phi, z | x)$. We assume the recognition model factorizes as follows under a mean-field assumption:

\begin{equation}
    q(\phi, z | x)=q_{\psi_\phi}(\phi | f(x)) q_{\psi_z}(z|x, \phi)
\end{equation}

The idea behind this factorization is that, we know from prior knowledge that the phase $\phi$ is well predicted by a physics informed hand-engineered transformation $f(x)$ \cite{zeng2024}. Specifically this transformation takes in composition and outputs eight relevant physical descriptors \cite{yan2021}; atomic size difference, mixing enthalpy, mixing entropy, Pauli electronegativity difference, molar volume, bulk modulus, melting temperature and valence electron concentration. In turn the neural network $\psi_\phi$ uses these engineered features to predict the binary phase. Note that $f(x)$ is a pre-specified transformation and not a learnable function. Finally the latent variable $z$ encodes everything else about the alloy $x$ conditioned upon values of phase. This recognition model allows us to encode hand-engineered features that we know are useful for predicting phase, and rely on the expressivity of neural networks to encode other information necessary for the eventual reconstruction of the training data through the decoder.

\subsubsection{Model training}
 We can train both the generative model and the recognition model simultaneously by maximizing the following variational objective function with respect to the neural network parameters for the generative and recognition model:
\begin{equation}
    \sum_{n=1}^N \mathcal{L}(\theta, \psi_z, x^n) + \gamma \sum_{m=1}^M \mathcal{L}^\textit{sup}(\theta, \psi_\phi, \psi_z; x^m, \phi^m)
\end{equation}

The first part of this objective function optimizes over all $n={1, \ldots, N}$ data points for which supervision is not available. This is the standard VAE evidence lower bound (ELBO) loss, which can generally be thought of as learning to reconstruct input data with some regularization on the latent space. The second term is the supervised loss and uses $M$ points where supervision is available i.e. points for which values for phase formation are available. Where the constant $\gamma$  is a hyperparameter that balances prediction accuracy for supervised learning and reconstruction accuracy for unsupervised learning. The objective above can then be approximated using a Monte Carlo estimator as explained in the survey by Luengo et al~\cite{luengo2020}. We utilize Pyro~\cite{bingham2018pyro}, a pytorch based probabilistic programming language, for model specification as well as training and inference.

Although all data points in the experimental dataset have labels, we split the data into three parts, and we treated them separately in the model training process.
The three parts include labelled data where the ground truth labels were used for training, unlabelled data where we only considered their reconstruction loss, and validation data that were used to validate predictions on both  reconstruction accuracy and single-phase formation probability.
The entire data set was randomly split into 864 labelled training examples, 296 unlabelled training examples, 75 validation examples, and 138 test examples.
Two hidden layers of size 100 each were used for both recognition and generative models. `Adam' optimizer was used for optimization of neural network parameters. An initial training rate of 10$^{-4}$ was employed with a 0.9 decay rate for momentum and 0.999 decay rate for squared gradients. The learning rate was reduced by a factor of 0.5 if no improvement in training has been observed for 200 consecutive steps. A batch size of 32 was used for training the model with up to 20000 epochs. The model with the highest validation accuracy was saved as the best model for hereafter inference and analysis. To examine the randomness of data splitting, five different random seeds were used to split the train, validation, unlabelled and test datasets. The mean and standard deviation of prediction accuracy and area under the curve for receiving operating characteristics are reported in Section~\ref{sec:class_results}.

\subsection{Post-hoc analysis: SHAP feature importance}
To analyze the impact of each feature on the classifier for the single-phase formation, we utilize the existing post-hoc analysis method implemented in SHAP~\cite{NIPS2017_7062}, a game theoretic approach to explain any model output.
Specifically, we used the model-agnostic kernel explainer. The training set was used as the background dataset to integrate out features, and the overall feature importance was evaluated on each instance of the test data.

\section{Results and Discussion}

\subsection{Classification and Reconstruction Performance}

\subsubsection{Classification for single phase formation}
\label{sec:class_results}

\begin{figure}[H]
    \centering
    \includegraphics[width=3.5in]{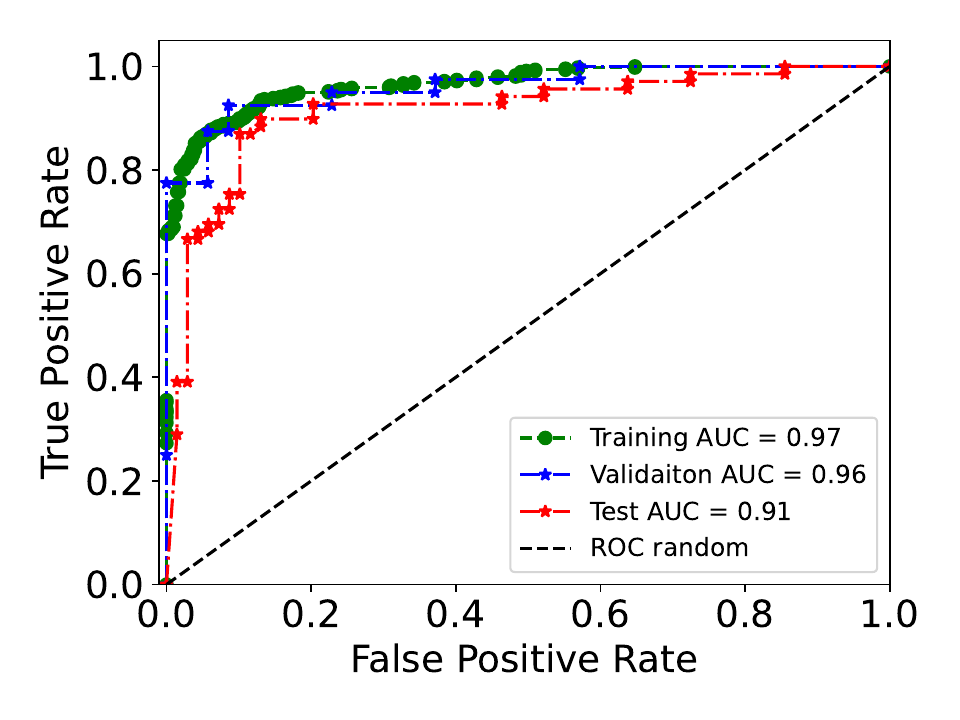}
    \caption{ROC curves for training, validation and test datasets.}
    \label{fig:roc}
\end{figure}

Using five random seeds for data splitting, the mean and standard deviation of prediction accuracy for training, validation and test are 0.883$\pm$0.027, 0.930$\pm$0.026 and 0.829$\pm$0.050, respectively.
The mean and standard deviation of area under curves (AUC) are 0.954$\pm$0.014, 0.955$\pm$0.029 and 0.890$\pm$0.025 for respective training, validation and test data. Figure \ref{fig:roc} shows the receiving operation characteristic (ROC) curves for training, validation and test datasets. The area under curves (AUC) are all no less than 0.91, suggesting a reliable prediction of single-phase formation.

\subsubsection{Alloy reconstruction}

\begin{figure}[H]
    \centering
    \includegraphics[width=6.5in]{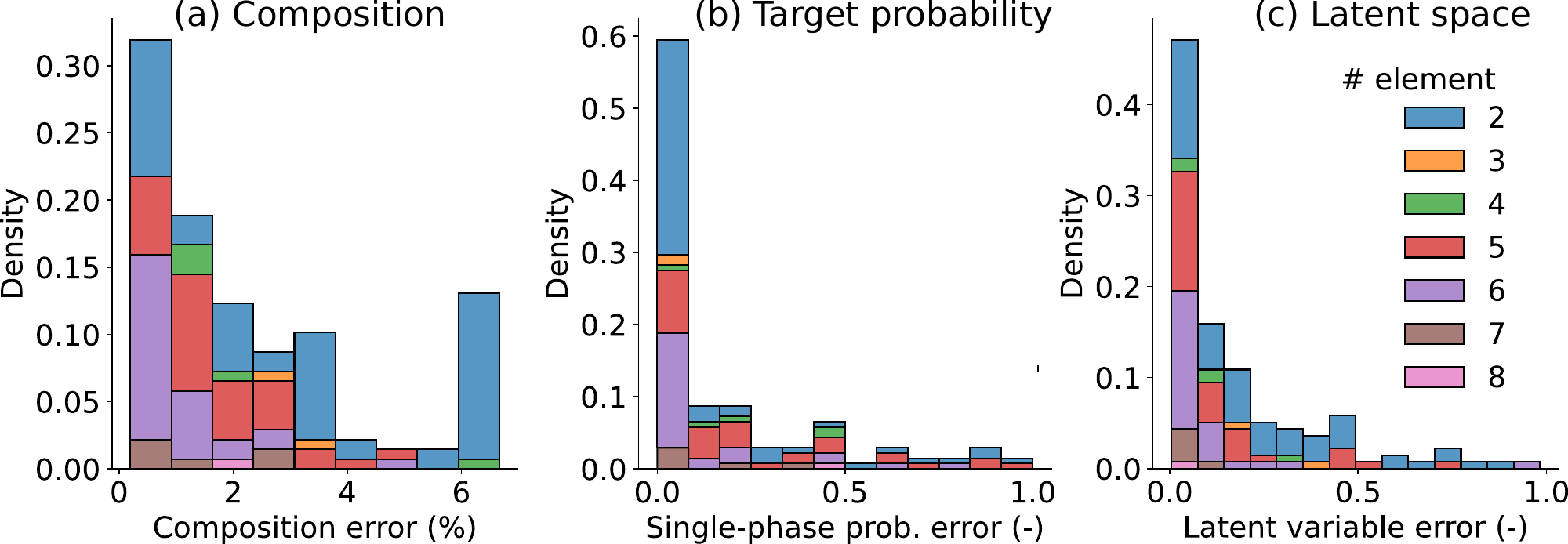}
    \caption{Comparison between original alloys and reconstructed alloys across 138 test data points: (a) Composition vectors, (b) Predicted single-phase probability, and (c) Latent variables.}
	\label{fig:alloy_reconstruct}
\end{figure}

We evaluated the recognition and generative models through an alloy reconstruction process.
We reconstructed alloys for all 138 data points in the test data, and compared the reconstructed alloys with the original alloys in terms of their composition vectors, predicted single-phase probability and latent variables. The results are shown in Figure~\ref{fig:alloy_reconstruct}. We observed a maximum MAE of ~$6\%$ for reconstruction of the composition vectors, with an average MAE of $2.3\%$. Higher error was observed in reconstructing binary alloys. MAEs for predicted single-phase probability and latent variables are concentrated at small values, with respective averages of 0.169 and 0.175. These results taken together show that the model was able to reliably capture enough information about the alloys to be able to reconstruct their composition and properties accurately.

Next, to demonstrate how the model can be used for deeper exploration, we chose a few HEA examples in the test dataset, and we compared the predicted class to the ground truth labels as well as the reconstructed alloys versus the original ones. We first compute the predicted probability of being single phase using the classifier head of the model. We then reconstruct the alloy given the composition features and the predicted probability using the generative model. The results are summarized in Table \ref{tab:vae_reconstruction}.
\begin{itemize}
    \item In the first example `\ce{Fe19Ni19Cr19Co13Al19Mo9}', We observe a predicted probability of $0.15$, which shows that the predicted class is highly likely to be multi-phase, consistent with the ground truth. The reconstructed alloy comprises of the same elements as the original alloy with slightly different composition (marginally higher Co and Ni compositions and marginally lower Cr and Al compositions).
    \item The second example `\ce{Al11Ti22V22Nb22Zr22}' is primarily made up of refractory elements with a small amount of Al and it forms a multiple-phase structure in experiments. With a predicted probability of $0.10$ the model also correctly predicts that this will form a multiple-phase structure. The reconstruction is again largely similar except with trace amounts of Cr and Ta which are not present in the original. While this isn't perfect reconstruction, note that Cr and Ta are in nature also refractory elements, similar to Ti, V, Nb and Zr.
    \item The third example `\ce{Al4Ti23Mo23V23Ta23}' is similar to the second example but with less Al. It tends to form a single-phase structure. The predicted class agrees well with the ground truth with a probability of $0.95$. reconstructed alloy contains three more elements (Cr, Nb and Zr) than the input alloy, which can be explained by the same rationale as for the second example, i.e. these additional elements are also refractory.
    \item In the last example, the predicted probability 0.52 of `\ce{Fe20Ni20Co20Ti20Cu20}' slightly favors the formation of single phase, which is indeed the ground truth. The reconstructed alloy also has more elements (Al, Cr, V, Mn) and no Ti. The replacement of Ti with Cr and V follows the same logic as the second and third examples. The addition of a small amount of Al may be attributed to the frequent appearance of Al in the training data while Mn, located in the middle of first-row transition metals, reconciles differences between different types of transition metals between (Fe, Co, Ni), (V, Cr) and Cu.
\end{itemize}

The above examples show that while the reconstructions are (expectedly) not exact, the ``errors'' in reconstruction happen along materially explainable directions - indicating that the latent space is encoding useful information about the alloy generation process.

\begin{table}[H]
\small
\centering
 \caption{Evaluation of semi-supervised autoencoder for alloy reconstruction. Since formulas for original alloys are varied, we converted all formulas to a standard format where compositions are rounded to integers with a sum close to 100, and element orders are sorted. Total compositions may not be exactly 100 because of the rounding issue.}
\begin{tabular}{ccccc}
\hline \hline
 Original alloy & Reconstructed alloy & Predicted probability & Ground truth label \\
\hline
\ce{Fe19Ni19Cr19Co13Al19Mo9} & \ce{Fe19Ni21Cr15Co21Al15Mo9} & 0.15 & 0 (Multiple phase) \\
\ce{Al11Ti22V22Nb22Zr22} & \ce{Cr3Al7Ti18V20Nb32Zr16Ta4} & 0.10 & 0 \\
\ce{Al4Ti23Mo23V23Ta23} & \ce{Cr1Al2Ti29Mo26V11Nb16Zr8Ta7} & 0.95 & 1 (Single phase)\\
\ce{Fe20Ni20Co20Ti20Cu20} & \ce{Fe17Ni23Cr13Co21Al2Cu5V10Mn10} & 0.52 & 1 \\
\hline \hline
\end{tabular}
\label{tab:vae_reconstruction}
\end{table}

\subsubsection{Data efficiency}

In comparison with supervised machine learning only, this proposed semi-supervised framework is more data efficient, and the reason is twofold.
First, the model learns a probabilistic feature--target relationship using prior distributions for features, target and latent variables. This expert-informed knowledge constrains model fitting and can be superior in generalizing the available data points while showing less variance when predicting on unknown data. Second, it leverages all relevant information from both labelled and unlabelled data. Although a classifier solely determines the target property once it is trained, the unsupervised learning is integrated into the supervised learning because the losses are added up and optimized together.
The priors also regularize the classifier head, and it enforces a more accurate description of feature--target relationship that influences the reconstruction process.

To verify the data efficiency of the proposed framework, we held out the same 138 data points for testing. The prediction accuracy was compared between semi-supervised learning and conventional supervised learning on the training and test data. Fully-connected dense neural network models were used for the supervised learning using hyper-parameters were optimized via a grid search and found to be close to ones chosen for the classifier of the semi-supervised learning, with an initial learning rate of 10$^{-4}$ and two 100-neuron hidden layers.
The results and code for hyperparameter tuning can be found in our open-source repository.
We varied the number of data points for each dataset to simulate situations where we have sufficient and limited amount of labelled data.

We summarized the prediction accuracy on the training and test data in Table \ref{tab:data_efficiency}. When there is more labelled data (first row), the semi-supervised learning performs similar to supervised learning in terms of test accuracy. Significantly, when the size of labelled data is much smaller (second row), the proposed semi-supervised model outperforms a fully supervised model in terms of test accuracy. It shows that the proposed semi-supervised method can perform as well as a fully supervised method when there is a lot of labelled training data available and outperforms fully supervised classification when fewer labelled data points are available. Note that the proposed model also learns a latent space that can be explored and used for generating compositions (aspects that are missing from a fully supervised model).
\begin{table}[H]
\small
\centering
 \caption{Data efficiency of the semi-supervised learning compared versus supervised learning only. `SS' and `SL' refers to respective semi-supervised learning and supervised learning.}
\begin{tabular}{ccccc}
\hline \hline
 Labelled data & Unlabelled data & Validation data & SS test accuracy & SL test accuracy\\
\hline
864 & 296 & 75 & 0.877 & 0.870  \\
247 & 790 & 198 & 0.841 & 0.826  \\
\hline \hline
\end{tabular}
\label{tab:data_efficiency}
\end{table}

\subsection{Latent space representation}

\subsubsection{Disentanglement from target property}

\begin{figure}[H]
    \centering
    \includegraphics[width=5.5in]{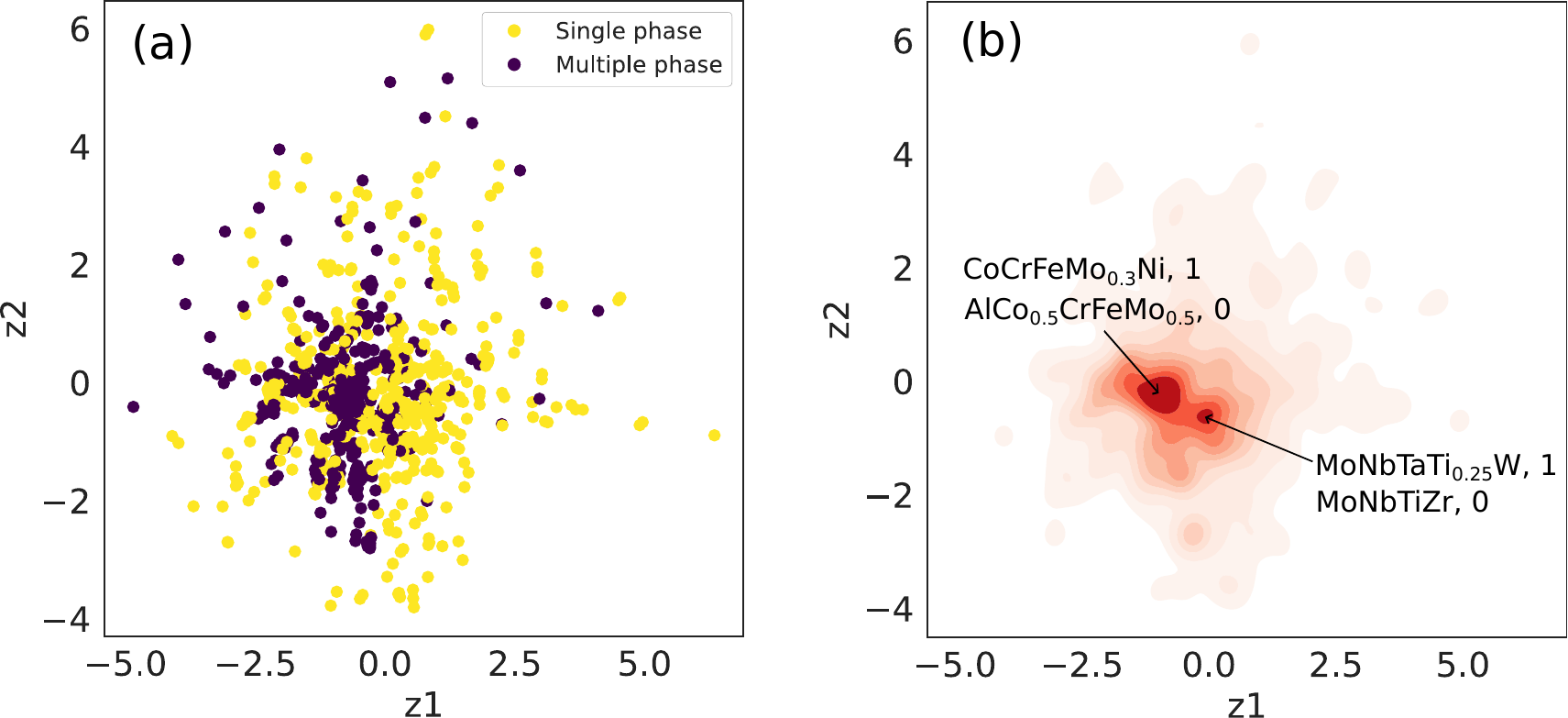}
    \caption{Disentangled latent representation: (a) Data colored by the target property, and (b) Data distribution shown as a kernel density plot. True labels for the HEAs are given as a binary presentation where `1' and `0' stand for single phase and multiple phase structures respectively.}
    \label{fig:latent}
\end{figure}

Latent space for all data points and their ground truth labels are shown in Figure \ref{fig:latent}(a), and a kernel density estimate for the data distribution is depicted in Figure \ref{fig:latent}(b). One can see that single phase alloys are mixed up with multiple phase alloys in the latent space, implying that the single phase formation aspect has been disentangled from this latent representation. Although the input alloys consist of up to 30 different elements and live in a high-dimensional space, more than 84\% data points are condensed into a small latent region ($z1 \in [-2.5, 2.5]$ and $z2 \in [-2, 2]$), as indicated by the kernel density plot in Figure \ref{fig:latent}(b). We also listed two pairs of high-entropy alloys in latent-space regions where data densities are the highest.
Each pair of points is of similar latent variables but has different true labels again showing that the space captures properties other than the phase formation.

\subsubsection{Association with other properties}

\begin{figure}[H]
    \centering
    \includegraphics[width=5.5in]{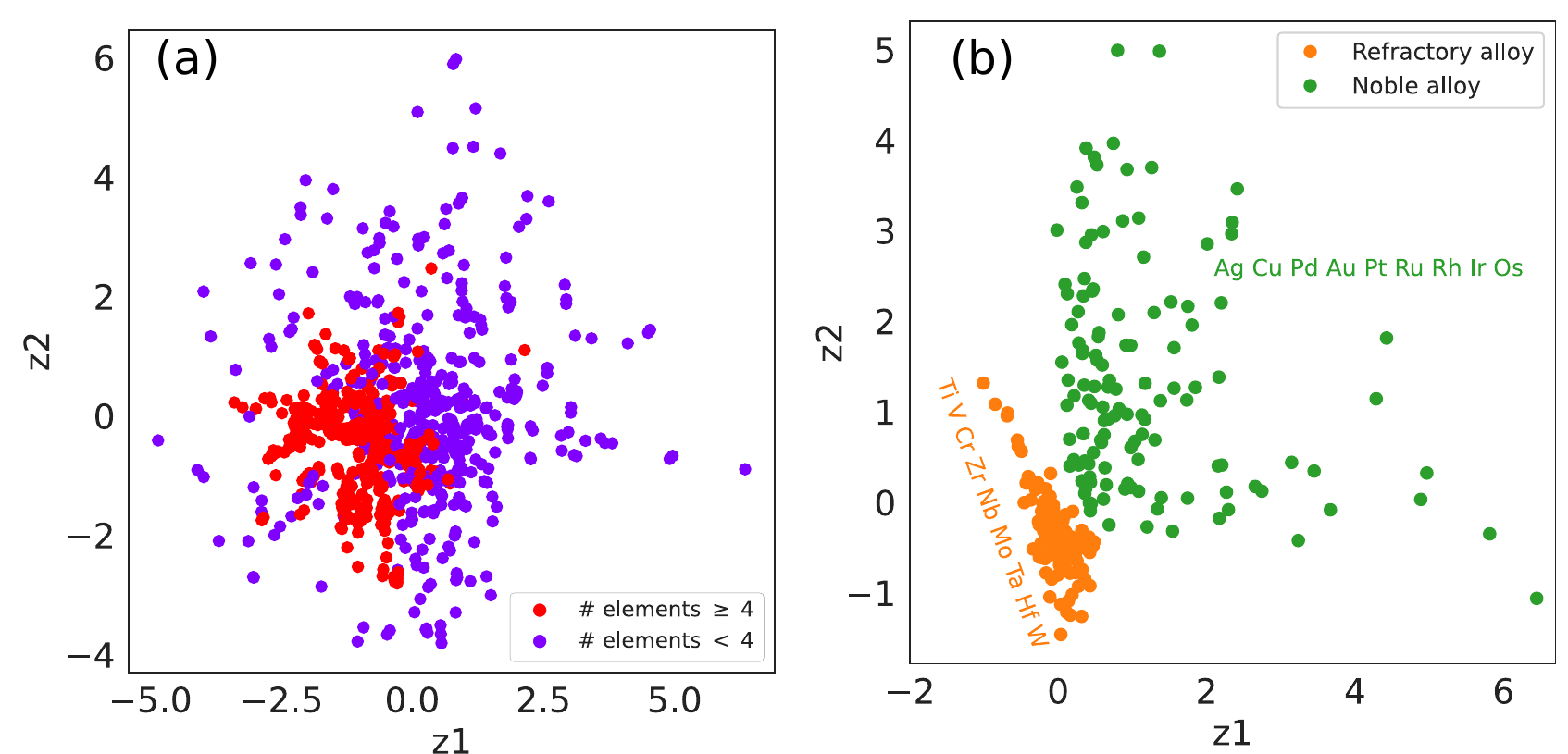}
    \caption{Association with other properties: (a) Data colored by number of elements, and (b) Latent variables for four-element HEAs generated from three groups of elements.  Element lists for each group are indicated by texts close to the corresponding point cloud.}
    \label{fig:latent_other}
\end{figure}

The proposed semi-supervised learning methodology allows us to disentangle the target property from the latent space, allowing the latent space to be implicitly associated with other properties. Figure \ref{fig:latent_other} shows the association of learned latent space with two other properties. Figure \ref{fig:latent_other}(a) represents the number of elements for each alloy in the full dataset. It is found that HEAs with no less than four elements are concentrated in a smaller region of the latent space compared to simpler alloys with less than four elements. In Figure \ref{fig:latent_other}(b), we created two groups of elements based on their positions in the chemical periodic table, including noble and refractory elements. One may argue that the exact group some elements belong to may vary by definition, but the purpose here is to group elements that at least share some similarity in atomic features. We then chose four elements from one group and created an equimolar four-element HEA, and we generated its latent representation. One can find that HEAs for each group of elements are located in distinct regions in the high data density region of the latent space (shown in Figure~\ref{fig:latent}). This shows that the model learned a general representation for elements across the chemical periodic table, which further explains why reconstructed alloys sometimes contain additional elements from the same group of elements and why high-dimensional composition features can be mapped into the more compact latent space.

\subsection{Inverse materials design}

The proposed semi-supervised variational autoencoder is useful for inverse materials design. With the classifier head in the recognition model, one can carry out high-throughput virtual screening of materials to identify potential candidates with desirable target properties. However, there is a chance that no desired materials can be found even if a wide range of composition space is scanned, making this method inefficient.

Another method is to start from the latent space and a given probability for single-phase formation. The generative model will create an alloy based on latent variables and single-phase formation probability. The generated alloy should however be re-examined by the classifier to ensure the predicted probability agrees with the initial input probability. For example, if we aim to find single-phase refractory HEAs, we first locate a latent point, such as $(z1, z2)=(0.0, -0.8)$, among the refractory alloy blob as shown in Figure \ref{fig:latent_other}.
For a formation probability of 0.9, the generated alloy is `\ce{Ti9V10Nb19Zr58Ta1Hf2}', and the classifier predicts it to be single phase with a probability close to 1. If we set the probability as 0.1, the alloy generated becomes `\ce{Cr3Al12Ti19V21Nb28Zr13Ta5}', and the predicted probability is 0.04. Although the learned latent representation generalizes many different elements, caution should be taken if we would like to generate alloys in regions where labelled data are scarce, which can result in high uncertainty for the predicted target property. As an example, if we start from  $(z1, z2)=(0.8, -0.5)$ which belongs to the magnetic alloy blob where data densities are low according to Figure \ref{fig:latent}(b). The generated alloy with an input probability of 0.1 is `\ce{Fe2Co81Ta14Pt2}', yet the predicted probability turns out to be 0.71.

\begin{figure}[H]
    \centering
    \includegraphics[width=5.5in]{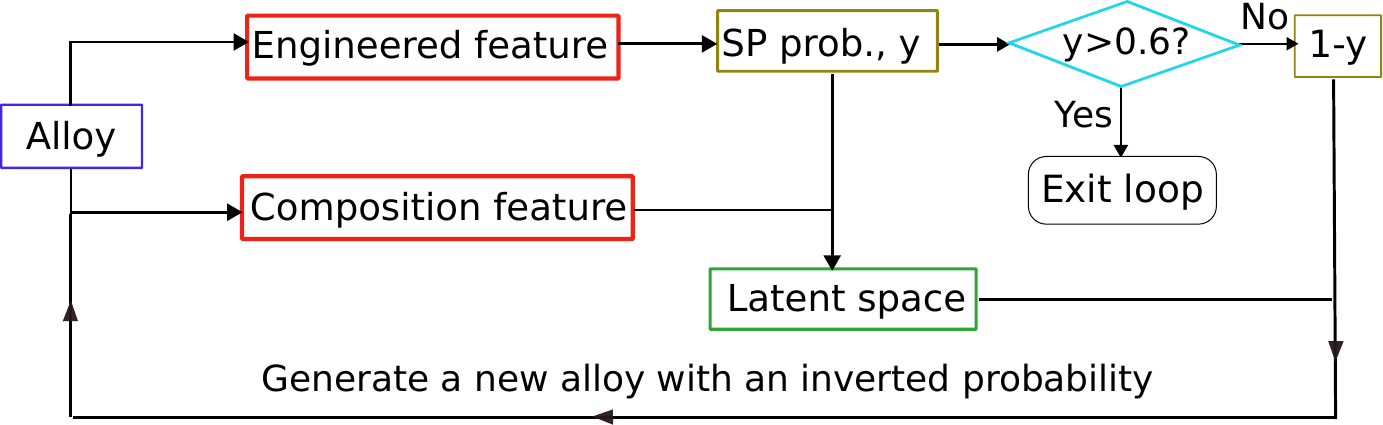}
    \caption{Workflow of the iterative process to search for single-phase alloys using the disentangled variational autoencoder. `SP' stands for single phase.}
    \label{fig:iterative_design}
\end{figure}

This inconsistency between input probability and predicted probability for alloys generated from low-density regions, motivates a third method. In practice, one may have a specific alloy in mind, but the alloy does not give the desired property. The goal is to nudge the alloy composition in directions where the desired property can be achieved without significant changes of alloy constituents. In this case, we can use an iterative process to find a single-phase alloy from an initial multi-phase alloy, as illustrated in Figure \ref{fig:iterative_design}. The initial alloy is converted to engineered features which are used to predict the single-phase probability. The composition features converted from the alloy are combined with the single-phase probability to generate its latent representation. We invert the single-phase probability, and the probability is concatenated with the latent representation to generate a new alloy. The loop ends when the single-phase probability predicted by the classifier is larger than a predefined cutoff (e.g. 0.6). We demonstrate this workflow using an initial alloy `Al$_{1.4}$Co$_{0.9}$Cr$_{1.4}$Cu$_{0.5}$Fe$_{0.9}$Ni$_1$' or equivalently `\ce{Fe14Ni16Cr22Co14Al22Cu8}'. The evolution towards a single-phase alloy follows the sequence:

\begin{equation*}\label{eq:phase_evolution}
\underset{y=0.12,\vec{z}=[-1.584 -0.201]}{\ce{Fe14Ni16Cr22Co14Al22Cu8}} \rightarrow \underset{y=0.48,\vec{z}=[-1.138 -0.163]}{\ce{Fe17Ni22Cr23Co25Al12}}
\rightarrow \underset{y=0.08,\vec{z}=[-0.985 -0.103]}{\ce{Fe19Ni22Cr19Co25Al15}} \rightarrow \underset{y=0.70,\vec{z}=[-0.846 -0.187]}{\ce{Fe21Ni22Cr22Co35}}
\end{equation*}

\noindent where the corresponding predicted single-phase probability $y$ and latent variables $\vec{z}$ are shown right below the chemical formula of the alloy.
One can see that after three iterations, the alloy eventually transforms to an alloy with a single-phase probability 0.70. In the first step, the alloy moves into a region without Cu, which enhances its single-phase probability. The second step detours around the target, transitioning to an alloy with reduced probability before it was finally inverted to a single-phase structure. The alloy evolution occurs in latent regions close to the initial point, hence it preserves many original elements or introduces new elements that are learned to be similar to existing ones. This probabilistic process offers more control over the inverse design process and provides intuitive interpretability of inversion of materials for desired properties. One can see that with principal elements of Fe, Ni, Cr and Co, it is unfavorable to form single-phase structures when Al and Cu are added. Moreover, the compositions of Al and Cu are almost evenly replaced by Fe, Ni and Co compositions while Cr compositions remain almost unchanged in the alloy evolution process. It suggests that Cr also reduces the probability of single-phase formation when mixed with the three magnetic elements. This observation agrees very well with our previous work, indicating that Al and Cr are adverse to the single-phase formation for high-entropy alloy AlCrNiCoFe~\cite{zeng2024}.

To systematically examine the interpolation capability of inverse materials design, we focus on a 2-D grid in the latent space, where $z1 \in [-0.1, 0.1]$ and $z2 \in [-0.5, -0.3]$. We generated a new alloy for each grid point using two different single-phase formation probability, including 0 (multi-phase), and 1 (single phase). In general, we found this region to be associated with refractory high-entropy alloys.
In Table~\ref{table:interpolation}, we summarize the results for a few examples. Values in the bracket indicate the predicted probability for the generated alloy. A full list of results for this systematic interpolation study could be found in our open-source repository.

\begin{table}[H]
\small
\centering
 \caption{Interpolation study for inverse materials design focused on a small latent region with high data densities. Values of $y$ are input single-phase probability used to generate the target alloys and values in the bracket are the predicted single-phase probability of generated alloys.}
\label{table:interpolation}
\begin{tabular}{ccc}
\hline
Alloy, $y$=1 & Alloy, $y$=0 \\
\hline
 \ce{Ti69Mo5Ta7W18} (0.001) & \ce{Cr18Al34Ti13Mo22V1Nb5Zr1Ta4Si2} (0.058) \\
 \ce{Ti52Mo11V1Nb2Ta15W19} (0.62) & \ce{Cr12Al19Ti21Mo26V3Nb8Zr3Ta6Si2} (0.007) \\
 \ce{Ti3Mo13V3Nb45Ta7W29} (0.84) & \ce{Cr5Al2Ti24Mo25V9Nb11Zr15Ta5Hf2Si3} (0.01) \\
\hline
\end{tabular}
\end{table}

At a probability 1 (targeting single-phase alloys), the generated alloys include only refractory elements including Ti, Mo, V, Nb, Ta and W. Switching the probability to 0 allows the model to transform a single-phase alloy to a multi-phase one. The results suggest that the addition of Cr, Al and Si shift a single-phase refractory alloy to be multi-phase, and the multi-phase alloy tends to have more elements than the single-phase counterparts.
Moreover, to validate the consistency between the input probability and single-phase probability of generated alloys, we used the classification head to predict the single-phase formation probability of generated alloys. We found that the predicted probabilities are in alignment with the input probability for all examples when input probability is 0 and most examples when input probability is 1.

\subsubsection{Post-hoc analysis for improved interpretability}

\begin{figure}[H]
    \centering
    \includegraphics[width=5.5in]{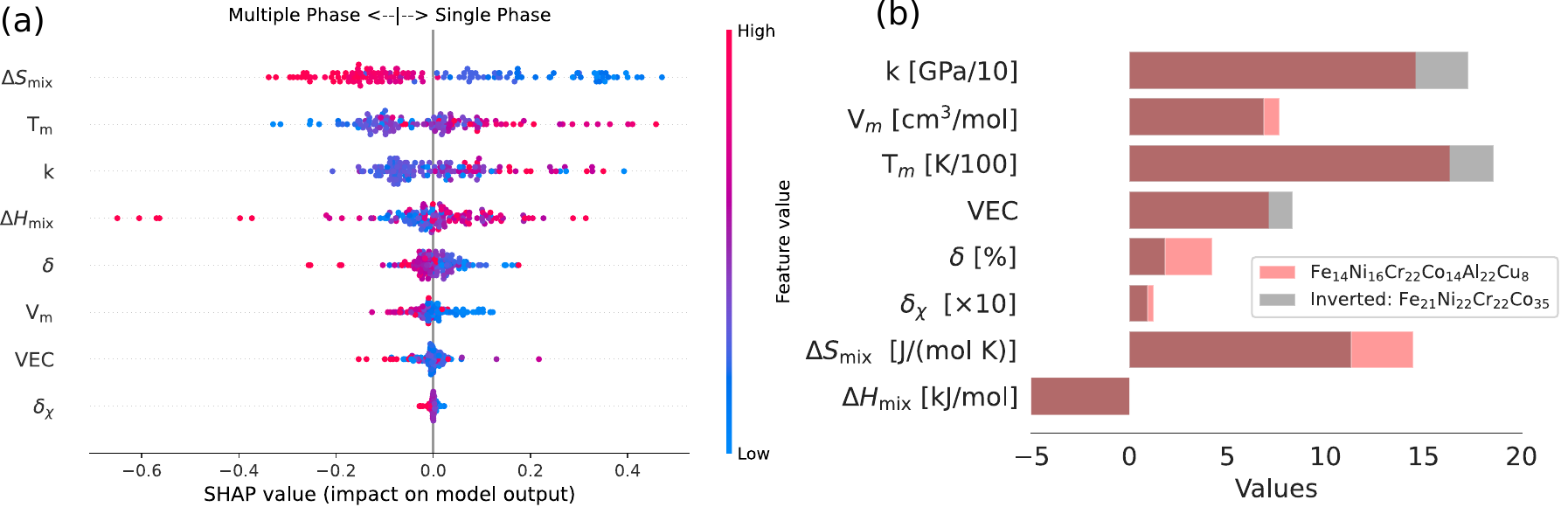}
    \caption{Post-hoc analyses for single-phase formation: (a) SHAP values for all data points, and (b) Feature values for an original multi-phase alloy and its inverted single-phase alloy.}
    \label{fig:shap}
\end{figure}

The semi-supervised learning approach is intended to be interpretable by design to the extent that it disentangles the latent space into the properties of interest. We can add additional one layer of interpretability by through post-hoc analysis methods. In the recognition model, the classifier head  takes as input the engineered features  and predicts the single-phase formation. That is still a black-box process. We utilize existing post-hoc explainability approaches, specifically Shapley values (SHAP) --- a widely used approach from cooperative game theory~\cite{NIPS2017_7062} --- to provide insight which input features can be attributed to the predictions made by the classification head for each input alloy. Figure \ref{fig:shap}(a) shows the aggregated feature importance values across all samples in the test dataset. One can see that lower mixing entropy and atomic size differences, and higher melting temperature and bulk modulus are more likely to form a single-phase alloy.Figure \ref{fig:shap}(b) compares the eight engineered features for an original multi-phase alloy `\ce{Fe14Ni16Cr22Co14Al22Cu8}' and the inverted single-phase alloy `\ce{Fe21Ni22Cr22Co35}'. It is clear that the inverted alloy is pushed toward the direction where it is of a much smaller atomic size difference, lower mixing entropy and higher melting temperature and bulk modulus, in satisfactory agreement with the feature importance analysis.

\section{Conclusion and Outlook}

We proposed a semi-supervised disentangled deep generative model for inverse materials design using probabilistic modeling. We demonstrated that the model can achieve better prediction accuracy than fully supervised machine learning when only a small amount of labeled data is available. The model is also interpretable by design as it disentangles the target property from other latent properties of the materials. These other properties can be explored using the learned latent space from the model. The model is also amenable to additional explainability as the target property prediction head acts as a classifier and can be investigated using post-hoc explainability methods.

We utilized an experimental high-entropy alloy dataset to demonstrate the feasibility of this approach. Specifically, we focused on the inverse design of high-entropy alloys that are likely to form single-phase structures. The model learned latent representations that were compact and disentangled from the target property, providing a design space with a separate and tunable target representation.
By using similar latent variables, this disentangled approach allows transformations across alloys with similar alloy constituents but distinct materials properties. We found that the latent space encoded other useful properties, including the number of elements and alloy types.
We demonstrated that using a well-trained disentangled variational autoencoder, inverse materials design can be conceived in three different ways, including high-throughput virtual screening using the classification/regression head, design from a latent-space region and an iterative design.

While a single target property is used in this work due to the data availability, this approach can be easily adapted for multiple properties as long as reasonable prior distributions for each property can be identified.
A case study is provided in the supporting information where one aims to find high-entropy alloys with desired phase formation, surface energies and stacking fault energies.
The approach implemented in this work may lead to alloys with harmful or expensive elements.
Hence, new methods to constrain material representations while searching for new materials are crucial to develop cost-effective materials with desired properties.
Uncertainty and inconsistency in the inverse materials design necessitates an active learning approach where reliable materials validation should be incorporated, and uncertainty estimate and retraining algorithms should be developed.
In this line of research, a human-in-the-loop intelligent interface may be beneficial to guide the search of new materials and identification of new data to improve the model, in particular for non-expert users.
We would also expect to see extensions of this approach to other types of materials representations (e.g. atomic structures, microstructures and crystal graphs etc) and models for inverse design (e.g. diffusion models and generative adversarial networks).

\subsection*{Data and Code availability}

Code and data used in this work can be found in a public Github repository:

\noindent \href{https://github.com/cengc13/d_vae_hea}{https://github.com/cengc13/d\_vae\_hea}.

\subsection*{Acknowledgments}

We are grateful to the valuable discussions with Andrew Neils, Mahsan Nourani and Jack Lesko.
Modeling in this work was carried out using the Discovery cluster, with assistance from the Research Computing team at Northeastern University. This work was carried out with financial support from the Roux Institute at Northeastern University and the Alfond Foundation.

\bibliographystyle{bibstyle}
\bibliography{references}

\end{document}